\documentclass[sigconf]{acmart}

\AtBeginDocument{%
  \providecommand\BibTeX{{%
    \normalfont B\kern-0.5em{\scshape i\kern-0.25em b}\kern-0.8em\TeX}}}

\setcopyright{acmcopyright}
\copyrightyear{2022}
\acmYear{2022}
\setcopyright{acmlicensed}\acmConference[GECCO '22 Companion]{Genetic and Evolutionary Computation Conference Companion}{July 9--13, 2022}{Boston, MA, USA}
\acmBooktitle{Genetic and Evolutionary Computation Conference Companion (GECCO '22 Companion), July 9--13, 2022, Boston, MA, USA}
\acmPrice{15.00}
\acmDOI{10.1145/3520304.3534026}
\acmISBN{978-1-4503-9268-6/22/07}

\usepackage[ruled]{algorithm2e}
\usepackage{multirow}
\usepackage{booktabs}
\usepackage{graphicx}

\newcommand{\eg}{{\em e.g.}}
\newcommand{\ie}{{\em i.e.}}

\begin{document}

%\title{Fast Lexicase Selection: Rethinking Weighted Shuffle for\\ More Effective Evaluation}
% \title{Going Faster and Hence Further with Lexicase Selection}
\title{Lexicase Selection at Scale}

\author{Li Ding}
% \authornote{}
% \orcid{}
\affiliation{%
  \institution{University of Massachusetts Amherst}
  \country{}
}
\email{liding@umass.edu}

\author{Ryan Boldi}
\affiliation{%
  \institution{University of Massachusetts Amherst}
  \country{}
}
\email{rbahlousbold@umass.edu}

\author{Thomas Helmuth}
\affiliation{%
  \institution{Hamilton College}
  \country{}
}
\email{thelmuth@hamilton.edu}

\author{Lee Spector}
\affiliation{%
  \institution{Amherst College}
  \institution{University of Massachusetts Amherst}
  \country{}
}
\email{lspector@amherst.edu}

\renewcommand{\shortauthors}{Ding, et al.}


\begin{abstract}
  Lexicase selection is a semantic-aware parent selection method, which assesses individual test cases in a randomly-shuffled data stream. It has demonstrated success in multiple research areas including genetic programming, genetic algorithms, and more recently symbolic regression and deep learning. One potential drawback of lexicase selection and its variants is that the selection procedure requires evaluating training cases in a single data stream, making it difficult to handle tasks where the evaluation is computationally heavy or the dataset is large-scale, \eg, deep learning. In this work, we investigate how the weighted shuffle methods can be employed to improve the efficiency of lexicase selection. We propose a novel method, fast lexicase selection, which incorporates lexicase selection and weighted shuffle with partial evaluation. Experiments on both classic genetic programming and deep learning tasks indicate that the proposed method can significantly reduce the number of evaluation steps needed for lexicase selection to select an individual, improving its efficiency while maintaining the performance.
\end{abstract}




\maketitle

\section{Introduction}

Genetic programming (GP) is a general methodology that has been widely embraced to tackle various kinds of problems including program synthesis, symbolic regression, statistical modeling, and machine learning. GP evolves programs by applying operations such as mutation and parent selection, mimicking the natural evolution processes, to the population of programs. Among many techniques that have been developed to determine the performance of individuals in parent selection, lexicase selection~\citep{helmuth2014solving,spector2012assessment} and its variants~\citep{helmuth2016lexicase,helmuth2015general,helmuth2014solving,liskowski2015comparison}
have been demonstrated to have more promising and robust performance than
others~\citep{fieldsend2015strength,galvan2013using,krawiec2015automatic} in a number of applications. The key idea of the lexicase selection is to gradually eliminate candidates by evaluating the population on each case in a randomly shuffled sequence of data. This procedure has been shown to bolster the diversity and generality in populations, which enhance the performance of evolved programs. 

Some recent work also shows that lexicase selection can be applied in
rule-based learning systems~\citep{aenugu2019lexicase}, symbolic
regression~\citep{la2016epsilon}, machine
learning~\citep{la2020learning,la2020genetic}, evolutionary
robotics~\citep{huizinga2018evolving,la2018behavioral}, and deep learning~\citep{ding2022optimizing}, helping enhance model performance and generalization. \citet{ding2022optimizing} propose gradient lexicase selection, a variant of lexicase selection that shows promising results on improving the generalization of deep neural networks (DNNs). Despite these recent successes, one potential drawback of the lexicase selection algorithm comes to our attention, especially for problems that involve computationally-heavy programs and large-scale datasets. Modern computing hardware, such as GPU clusters, have moved towards single-instruction multiple-data (SIMD) architectures. Since lexicase selection operates in a sequential single-data-stream fashion, it hardly can be optimized with parallel computing to reduce its running time. For example, in \citet{ding2022optimizing}, each lexicase selection process involves running inference of a deep neural network on at most 50,000 training cases per generation, with a batch size of just $1$. 

In this work, we aim to explore ways to speed up lexicase selection while preserving its core behavior of performing sequential evaluation of training cases for selection. An essential observation of the lexicase selection algorithm is that the training cases are considered in randomly-shuffled ordering, with cases earlier in the ordering receiving more attention than those later in the ordering. It is straightforward to presume that if the early training cases have more selection pressure, the selection process may be terminated earlier with a reduced number of evaluations on training cases. With this intuition, we revisit the weighted shuffle methods~\cite{troise2018lexicase} that shuffle the training cases in a non-uniform manner for lexicase selection, with the goal of finding a weighted shuffle that does not negatively affect problem-solving performance but can improve the average running time of lexicase selection.

While \citet{troise2018lexicase} did extensive experiments and comparisons of different shuffling methods, a prominent missing piece is that they only investigate if weighted shuffling has an effect on the final performance, but not the algorithm running time. There were two potential reasons: 1) the computation needs for classic GP problems are usually small, so the common implementation of lexicase selection assumes that the population has already been evaluated by all the training cases, in which case the selection is performed on the error vectors; 2) the weight assignment for weighted shuffling is based on those error vectors produced by a complete evaluation. However, when using lexicase selection for problems with computationally-heavy evaluation process and/or large-scale datasets, \eg, deep learning problems, it is obvious to see that such evaluate-then-select paradigm is no longer appropriate as it will introduce a lot of redundancy in computation.

Our goal is to reduce the number of evaluations needed for selection, in which case only the evaluations that are necessary for the selection process will be performed. This also means that only partial evaluation of the training set will be available for assigning weights to each training case for the weighted shuffling. To solve the above issues, we propose Fast Lexicase Selection: a new variant of lexicase selection method using weighted shuffle with partial evaluation. Our method performs evaluation and selection simultaneously. It can also assign and update weights for shuffling training cases based on evaluating only a portion of the training cases. We also extend the idea of our method to other variants of lexicase selection, \eg, gradient lexicase selection~\citep{ding2022optimizing}, to target on problems such as deep learning.

For experiments, we test the proposed algorithm on both the classic GP problems and the benchmark deep learning problem (image classification), where the latter involves large-scale datasets and heavy computation. Empirical results show that the proposed algorithm has a significant improvement in efficiency with a reduced number of evaluations, while maintaining similar performance to the regular lexicase selection.

\section{Background and Related Work}

\subsection{Preliminaries of Lexicase Selection}

Lexicase selection considers performance on individual training cases as opposed to aggregated fitness or accuracy metrics~\cite{helmuth2014solving,spector2012assessment}. This has the benefit of including some semantic information about the performance of the population to potentially help guide search. Due to the lack of aggregation, lexicase selection tends to select specialist individuals that trade being somewhat good at all cases for being elite on a subset of them~\cite{helmuth2019lexicase,helmuth2020importance}. This tends to have beneficial effects on behavioral diversity~\cite{helmuth2016lexicase,moore2018tiebreaks}, and success rates when compared to tournament and fitness-proportionate selection~\cite{helmuth2014solving}. The most recent work~\citep{ding2022optimizing} using this selection method proposed an evolutionary framework, gradient lexicase selection, that combines stochastic gradient descent and lexicase selection to improve the generalization of deep neural networks.

For every selection event, the training cases are shuffled into a random order that will be used to select one parent. Using this random case ordering, lexicase selection filters down the population by iteratively removing all individuals that do not exhibit exactly the best performance on each case, in the order they are presented. When a single individual remains, it is returned as the selected parent. If all the cases have been traversed and there are still individuals in the population, a random individual from the remaining group is returned.

One potential issue with lexicase selection and its variants in general, however, is that the selection process operates in a sequential single-data-stream fashion. In other words, the algorithm's time complexity can not be efficiently reduced by using modern parallel computing architectures. When the training cases for selection are large scale and the evaluation of each training case is computationally expensive, lexicase selection may have significant disadvantages when it comes to computation. In this work, our goal is to improve the running time of lexicase selection while maintaining its core behavior of looking at each individual case instead of using aggregated fitness metrics.

It is important to note that in lexicase selection, the cases are shuffled between selection events, so different cases are given different priority over the course of the selection process. In general, training cases that show up earlier have more impact on the selection process, since they are responsible for filtering out the most individuals from the population. It is likely that cases that come at the end have little to no effect as the selected individual could have been returned before these cases are ever seen.

\subsection{Improving Selection Efficiency}

% While not directly regarding lexicase selection, there is a large historical literature surrounding tweaking selection methods to improve efficiency in GP.

There is a large historical literature surrounding tweaking selection methods to improve efficiency in GP. \citet{ga94aGathercole} introduce Dynamic Subset Selection, where subsets of the training data that are harder are sampled and used for selection in tournament selection. 
% They found that this resulted in better results in less than 20\% of the time.
\citet{poliBackwardsGP2005} ran GP backwards, only evaluating individuals that would eventually be selected to be part of a tournament selection pool. More recently, \citet{chitty2018Paralleltourn} harnesses the power of parallel computing in order to speed up tournament selection. These methods have shown great strengths in increasing the efficiency of selection methods in GP.

More specifically for lexicase selection, \citet{DeMelo2019LexSpeed} introduced Batch Tournament Selection (BTS), a hybrid of tournament and lexicase selection, which attempt to incorporate the idea of lexicase selection into tournament selection to improve both efficiency and quality of solutions. ~\citet{aenugu2019lexicase} proposed Batch Lexicase Selection, which is a variant of lexicase selection that considers batched data during selection events. However, \citet{ding2022optimizing} find that batch lexicase selection does not perform as well as lexicase selection on some large-scale problems, such as image classification. In this work, our method aims to directly increase the efficiency of lexicase selection, which can be further extended to the above hybrid methods.

\subsection{Weighted Shuffle}

% \todo{Ryan}{try to summarize the related parts that has been explored in this work from GPTP Sec. 3, with references from GPTP Sec. 2)}

Lexicase selection with weighted shuffle~\cite{troise2018lexicase} is an attempt to change the way cases are shuffled every selection event. Instead of uniform random shuffling, weighted shuffling techniques bias the resultant ordering of cases based on a given metric. In this work, we will be considering only the weighted and ranked shuffling methods, with the Number-of-Zeros (easy first) and Number-of-Nonzeros (hard first) bias metrics.

\paragraph{\textbf{Weighted shuffle}}
Weighted shuffle assigns a weight to each training case based on the chosen bias metric. When selection requires a shuffled list of cases to be built, cases are chosen iteratively where cases with higher weights have a higher chance of being selected. This case shuffling technique could be thought of as a roulette wheel, where cases with a higher weight have a larger slice of the roulette wheel. Every time we need to pick a case to add to the list, we update the case probabilities and re-roll the wheel.  This entire process is repeated for every selection event in a generation, using the same case weights at the beginning. 

\paragraph{\textbf{Ranked shuffle}}
Ranked shuffle first ranks the cases based on the chosen bias metric, and sorts the cases using this ranking. Then, an upper bound is randomly selected between 1 and the number of training cases, inclusive. The selected case index is then chosen randomly from 1 to the upper bound, inclusive. The case at this selected index is added as the first case in the shuffled case list. This process is repeated for the remaining cases, building up the shuffled case ordering. Ranked shuffle is likely to place cases with low rank first in the final case ordering as these low ranks are likely to be between 1 and the selected upper bound. This method differs from Weighted shuffle due to it ignoring the magnitude of differences in weight between cases.

\paragraph{\textbf{Bias Metrics}}
The bias metrics used in the weighted shuffle selection process inform the shuffling method of how to assign weights to each training case. Past implementations of weighted shuffle have had access to how every individual in the population performed on each training case before selection begins. Therefore, this information can be utilized in the selection process as follows. The Number-of-Zeros bias metric counts the number of individuals that achieve zero error on a given training case. Shuffling using this metric would result in a bias towards easier cases earlier in the case ordering as easy cases likely have more individuals achieving perfect scores on them. The Number-of-Nonzeros bias metric counts the number of individuals that do not have zero error on a given training case. This metric would bias the case ordering towards placing harder cases earlier in the shuffle. For the applications studied in this work, we propose a general system to update these case weights \emph{without} evaluating every individual on every case (as this process might be computationally intensive when performing large-scale evolutionary optimization.\\

While \citet{troise2018lexicase} compared these weighted shuffling methods by referring to their solution rates and behavioral diversity, an empirical study comparing these methods with respect to efficiency was not performed.

\section{Method}

In this section, we first introduce a new method, weighted shuffle with partial
evaluation, that can assign and update weights of training cases with a
partially-observed performance measure. We then apply this method to lexicase
selection and propose two new variants of the algorithm, fast lexicase selection
and fast gradient lexicase selection, for which the computation cost may be
effectively reduced by having less training cases required to be evaluated.

\subsection{Weighted Shuffle with Partial Evaluation}

Our goal is to harness the benefit of weighted shuffle to reduce the number of
evaluations needed to be performed on training cases. In our case study, and the work performed by~\cite{troise2018lexicase}, the weighted shuffle method requires
all the training cases to be evaluated before the lexicase selection process, which
is not actually reducing the number of individual program evaluations in practice.

\begin{algorithm}[t]
    \KwData{
    \begin{itemize}
        \item \texttt{cases} - a sequence of all the data samples to be used in selection with default ordering
        \item \texttt{candidates} - the entire population of programs
        \item $W$ - a weight vector for all the \texttt{cases}
    \end{itemize}
    }
    \KwResult{
    \begin{itemize}
        \item an individual program to be used as a parent
        \item an updated weight vector $W$
    \end{itemize}
    }
    
    \texttt{shuffled\_cases} $\gets$ Weighted\_Shuffle(\texttt{cases}, $W$)

    \For{\texttt{case} in \texttt{shuffled\_cases}}{
        $i \gets$ the index of $case$ in \texttt{cases}

        \texttt{results} $\gets$ evaluate \texttt{candidates} on \texttt{case}

        $W[i] \gets$ Bias\_Metric(\texttt{results})

        \texttt{candidates} $\gets$ the subset of the current \texttt{candidates} that have exactly best performance on \texttt{case}

        \If{\texttt{candidates} contains only one single \texttt{candidate}}{\KwRet{\texttt{candidate}, $W$}}
    }
    \texttt{candidate} $\gets$ a randomly selected individual in \texttt{candidates}

    \KwRet{\texttt{candidate}, $W$}

    \caption{Fast Lexicase Selection for Parent Selection}
    \label{alg:flexi}
\end{algorithm}

The weighted shuffle method scores each training case based on some bias metric that
summarizes the performance of the population on that specific case. In this
work, we consider the scenario that not all the training cases are evaluated during
selection at each generation. We propose a novel method, weighted shuffle with partial evaluation, to handle such a situation. 

We begin by initializing a weight vector $W = [w_1,w_2, \cdots, w_n]$ as the weights assigned to each of the $n$ training cases, with some pre-defined bounds
$w_i \in [w_{min}, w_{max}], i=1,2,\cdots, n$. With a larger value of $w$, the
training case is more likely to appear earlier when shuffled.\footnote{The shuffling method used in this work is the standard weighted sampling from the multinomial probability distribution, as defined by the weights. We use the PyTorch implementation of the function \texttt{torch.utils.data.WeightedRandomSampler}.} The value of $w_{min}$ and $w_{max}$ depends on the bias metrics one choose to use. In general, we assume that $w_{min}>0$ to ensure that easy cases still have possibility to come early in the ordering.

Since none of the training cases have been evaluated prior to the selection event, we propose two options for assigning the
initial weights: 1) Default-Min: let $w_1 = w_2 = \cdots = w_{min}$, such that
the unevaluated cases are default to be least likely to appear early among all
the cases; 2) Default-Max: $w_1 = w_2 = \cdots = w_{max}$, meaning the
unevaluated cases are most likely to appear early in the shuffled queue of training
cases. 

For each generation, we first perform weighted shuffle on the sequence of training
cases based on the current weight vector $W$. Let $p^*$ denote the population
size and $p\in [0,p^*]$ to be the size of current population (after potential
removal of some offspring by the ongoing selection event). During the selection
process, after each evaluation step on training case $i$, we update its assigned
weight $w_i$ according to the performance of the current population, by using some bias metrics.

\begin{algorithm}[t]
  \KwData{
  \begin{itemize}
    \item \texttt{data} - the whole training dataset in a default order
    \item \texttt{candidates} - set of $p$ instances of the DNN model
    \item $W$ - a weight vector for all the samples in \texttt{data}
  \end{itemize}
  }
  \KwResult{
  \begin{itemize}
    \item an optimized DNN model
  \end{itemize}
  }
  
  \tcp{K training epochs}

  set $w_{min}=1$ and $w_{max}=p+1$

  initialize $W$ with $w_{min}$ (Default-Min) or $w_{max}$ (Default-Max)
  
  \For{$epoch = 1:K$}{
    \texttt{subsets} $\gets$ $p$ equal-size subsets obtained through random sampling from the entire \texttt{data} without replacement

    apply gradient descent and backpropagation to optimize each of the $p$ \texttt{candidates} on each of the $p$ \texttt{subsets} respectively

    \texttt{shuffled\_cases} $\gets$ Weighted\_Shuffle(\texttt{data}, $W$)

    \texttt{parent} $\gets$ \textit{None}

    \For{\texttt{case} in \texttt{shuffled\_cases}}{
      $i \gets$ the index of $case$ in \texttt{data}

      \texttt{results} $\gets$ evaluate \texttt{candidates} on \texttt{case}

      $W[i] \gets$ Bias\_Metric(\texttt{results})

      \texttt{candidates} $\gets$ the subset of \texttt{candidates} that have exactly best performance on \texttt{case}

      \If{\texttt{candidates} contains only one single \texttt{candidate}}{
        \texttt{parent} $\gets$ \texttt{candidate}
        break
      }
    }

    \If{\texttt{parent} is \textit{None}}{
        \texttt{parent} $\gets$ a randomly selected individual in \texttt{candidates}
    }

    \texttt{candidates} $\gets$ set of $p$ instances of the DNN model copied with the same parameters as \texttt{parent}

  }
  \Return{parent}
  \caption{Fast Gradient Lexicase Selection for Optimizing Deep Neural Networks}
  \label{alg:fglexi}
\end{algorithm}

\subsection{Fast Lexicase Selection}

By introducing the proposed weighted shuffle with partial evaluation method into
the algorithm pipeline, we propose two variants of lexicase selection,
\textit{Fast Lexicase Selection} and \textit{Fast Gradient Lexicase Selection}. 

\paragraph{\textbf{Fast Lexicase Selection.}}
We propose a general methodology for parent selection that combines
lexicase selection with weighted shuffle and online weight updating, named Fast lexicase selection. The algorithm is outlined in Alg.~\ref{alg:flexi}. 

The key idea of fast lexicase selection is
using weighted shuffle rather than random shuffle to avoid unnecessary
evaluation on training cases that do not result in effective selection of cases. For
example, if all the candidates succeed on an easy training case, it is more likely
that in the next generation they will again all succeed. Such a case has no
selection pressure on the population, and thus shall be avoided for better
efficiency. Especially for lexicase selection that operates in a
single-data-stream fashion, if training cases with higher selection pressure can be
prioritized in the training sequence, the computation cost may be effectively
reduced by having less training cases needed to be evaluated for each generation. 

\paragraph{\textbf{Fast Gradient Lexicase Selection.}}
In this work, our goal is to reduce the runtime of lexicase selection on
large-scale computation problems. To better illustrate this point, we further
extend the idea of fast lexicase selection to the gradient lexicase selection
method, which is a variant of lexicase selection that focuses on optimizing deep
neural networks. An outline of fast gradient lexicase selection is shown in Alg.~\ref{alg:fglexi}.

\section{Experiments on GP Problems}

In order to compare the efficiency of the above selection methods in a well established field, we performed GP runs evolving solutions to two different program synthesis problems. We introduce a new measure to evaluate the efficiency of said selection strategies and also verify that the relative success rate results match those found by \citet{troise2018lexicase}. Outlined below is the experimental setup and the results from these experiments.

\subsection{Experimental Setup}
\paragraph{\textbf{Problems}}
The two problems chosen for this investigation come from the first General Program Synthesis Benchmark Suite~\cite{helmuth2015general}. This benchmark suite contains programming problems that deal with a range of data types and programming constructs. The two problems chosen from this benchmark suite are \emph{Mirror Image} and \emph{Last Index of Zero}. Solution programs to \emph{Mirror Image} receive two vectors of integers, and return a Boolean indicating whether the vectors are the reverse of each other. For \emph{Last Index of Zero}, solutions must take as input a vector of integers (one of which is a zero), and return the index of the last zero in the vector. The datasets we used consist of input and output cases for these problems, as outlined in the benchmark suite~\cite{helmuth2015general}.

%The final problem, which we call \emph{Simple Regression}, calls for programs to emulate the function $f(x) = x^3 - 2x^2 - x$. Solutions to this problem both take as input, and output integer values.

\paragraph{\textbf{PushGP}}
We use the PushGP system to evolve solutions to the problems described above. PushGP is a GP system that evolves programs written in the Push programming language~\cite{spector2001autoconstructive,spector2002genetic}. The Push programming language is a stack-based language that facilitates the use of multiple data types and complex programming paradigms such as iteration and recursion~\cite{spector2005push3}.  We used an implementation of PushGP, Clojush\footnote{https://github.com/lspector/clojush}, that is written in Clojure.  Table~\ref{tab:pushGPparam1} outlines the system parameters used for the \emph{Mirror Image} and \emph{Last Index of Zero} problems.

\begin{table}[]
    \centering
    \begin{tabular}{lr}
    \toprule 
    \textbf{Parameter}     & \textbf{Value} \\
    \midrule
    runs per problem & 50 \\
    population size & 1000 \\
    maximum generations & 300 \\
    \midrule
    \textbf{Genetic Operator} & \textbf{Probability} \\
    \midrule
    alteration & 0.2 \\
    uniform mutation & 0.2 \\
    uniform close mutation & 0.3 \\
    uniform mutation followed by alternation & 0.5 \\
    \bottomrule
    \end{tabular}
    \caption{PushGP system parameters for the \emph{Mirror Image} and \emph{Last Index of Zero} problems. These parameters were picked as a rule of thumb to better compare to prior work.}
    \label{tab:pushGPparam1}
\end{table}

\paragraph{\textbf{Measures}}
We propose a new measure, the \emph{average number of evaluations}, in attempt to measure the efficiency of the various selection methods discussed above. This metric is a rough measure of the computational effort required to perform the lexicase selection procedure to select individuals. Instead of using wall-clock times like explored in previous selection strategy comparisons \cite{DeMelo2019LexSpeed, ga94aGathercole, poliBackwardsGP2005}, we use a measure that provides a hardware-agnostic efficiency comparison. We define an evaluation as the act of either computing the error for an individual (by evaluating the Push program they produce) on a specific case, or looking up its error on this case from a different selection. To calculate the number of evaluations for a single selection event, we add together the number of individuals remaining after every case is visited. For example, if there are $10$ individuals in the population, and $3$ remain after visiting the first case, and $1$ remains (and is selected) after visiting the second case, then the number of evaluations for this selection event is $10 + 3 = 13$. This metric represents the number of times (in an isolated selection event) a new individual would have to be evaluated on a new case from scratch.

It is important to note that the selections we are performing in our GP runs are \emph{not} isolated selection events as we can (and do) reuse evaluations of an individual on a specific training case from other selection events in the same generation to help increase performance. This is because it is sometimes the case in lexicase selection that one individual is evaluated on the same training case twice or more in a generation during the selection of different parents. Furthermore, the PushGP implementation we are using evaluates \emph{all} individuals on \emph{all} training cases at the beginning of every generation, for use in all selection events. Nevertheless, we believe that measuring the average number of evaluations is still a meaningful metric that can be broadly used to compare the efficiency of selection procedures. For different implementations of PushGP, or uses of lexicase selection outside of GP like those described in this paper and beyond, this metric can be used as a much closer correlate of run time. The lower the average number of evaluations, the more efficient the selection procedure is likely to be. This measure can also be thought of as how effective the cases are at filtering down the population size to more computationally manageable numbers. 

To calculate the \emph{number of evaluations} for a given generation, we sum together the number of evaluations performed in every selection event in that generation. For our GP runs, we average these values across all of the runs that are still active at this generation. This means that runs finishing early will not deceptively decrease the number of comparisons occurring during that generation.
%We could also think of this as normalizing the graphs to success rate, 
Along with this, we will also measure whether the various weighted shuffling methods in fast lexicase selection are comparable to lexicase selection in terms of \emph{success rate}. The number of successes in the following experiments refers to the number of runs (out of the 50 total) that successfully evolve a solution with perfect scores on all the held out training cases.

\begin{table}[]
    \centering
    \begin{tabular}{lcc}
        \toprule
        \multirow{2}{*}{\textbf{Selection Method}}& \multicolumn{2}{c}{\textbf{Success Rate (/50 Runs)}} \\
         & \emph{Mirror Image} & \emph{Last Index of Zero}  \\
        \midrule
        Lexicase & 43 & 20 \\
        Weighted - Num-Nonzeros & 45 & 11 \\
        Ranked - Num-Nonzeros & 46 & 21 \\
        Weighted - Num-Zeros & 48 & \underline{\textbf{2}}\\
        Ranked - Num-Zeros & 35 & \underline{\textbf{5}} \\
        \bottomrule
    \end{tabular}
    \caption{Success Rates of GP runs on the \emph{Mirror Image} and \emph{Last Index of Zero} problems, utilizing a variety of selection techniques. Underlined are the results that were significantly worse than lexicase selection. No results were significantly better than lexicase selection across either problem.}
    \label{tab:GP_Success}
\end{table}

\subsection{Results}  \label{sec:caseresult}

Table~\ref{tab:GP_Success} presents the differences in success rate across these selection methods. None of the weighted or ranked shuffling methods performed significantly better than lexicase selection. Also, none of the hard-first metrics performed significantly worse than lexicase selection. These findings align with those found by \citet{troise2018lexicase}.
% Maybe more to say about these results

\begin{figure}
    \centering
    \includegraphics[width=0.99\linewidth]{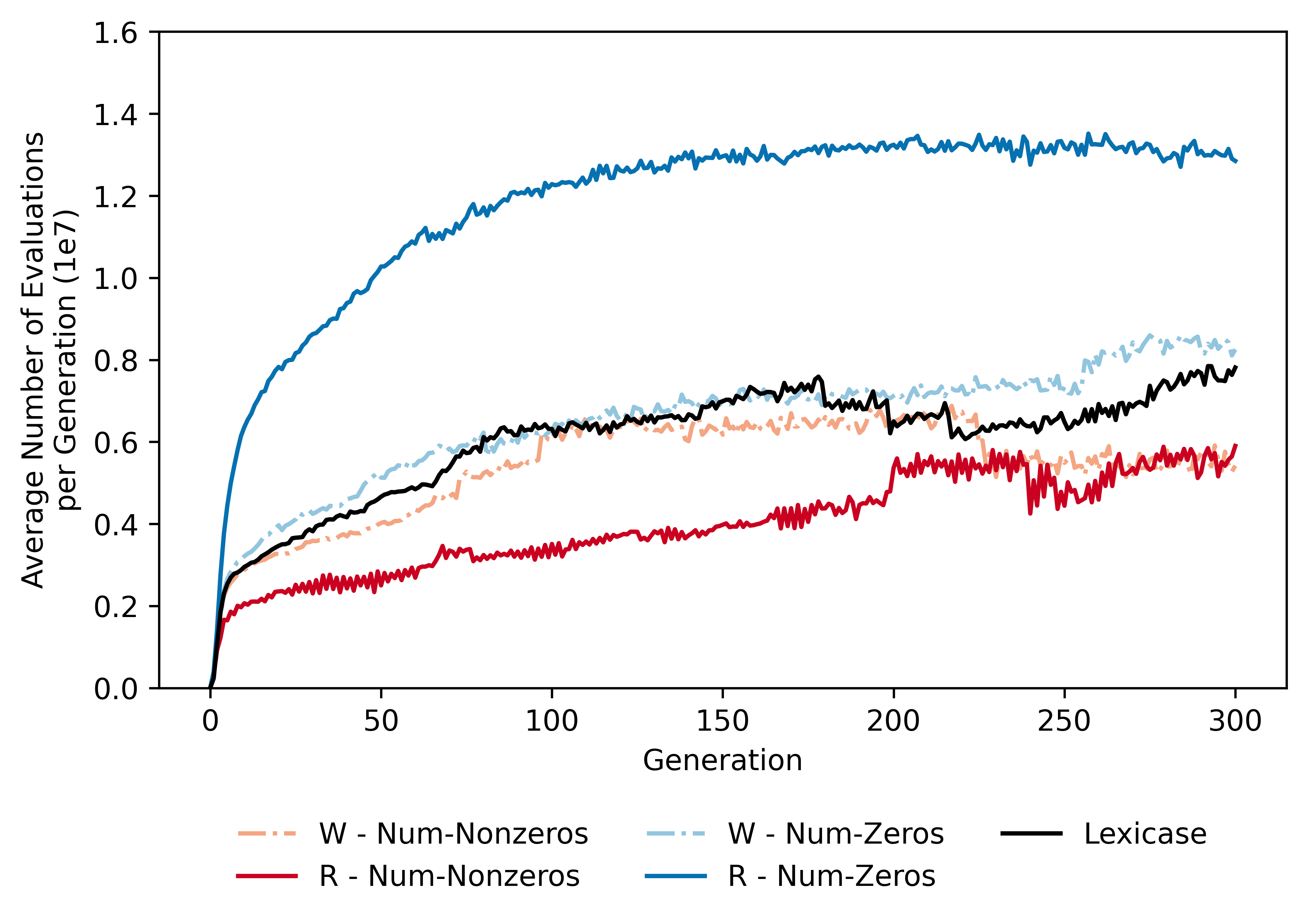}
    \caption{Average number of evaluations performed in a given active generation over evolutionary time while solving the \emph{Mirror Image} problem. The hard first metrics (those using the Number-of-Nonzeros bias metric) are shown in red and the easy first metrics (Number-of-Zeros) are shown in blue. The weighted shuffling technique is presented as dotted lines, and the ranked shuffling technique is shown in solid lines. Lexicase selection is shown in black. Ranked shuffling using the Number-of-Nonzeros bias metric consistently performs less comparisons per generation than all other shuffling methods, including lexicase selection (random shuffling).}
    \label{fig:MI_COMP_NC}
\end{figure}

\begin{figure}
    \centering
    \includegraphics[width=0.99\linewidth]{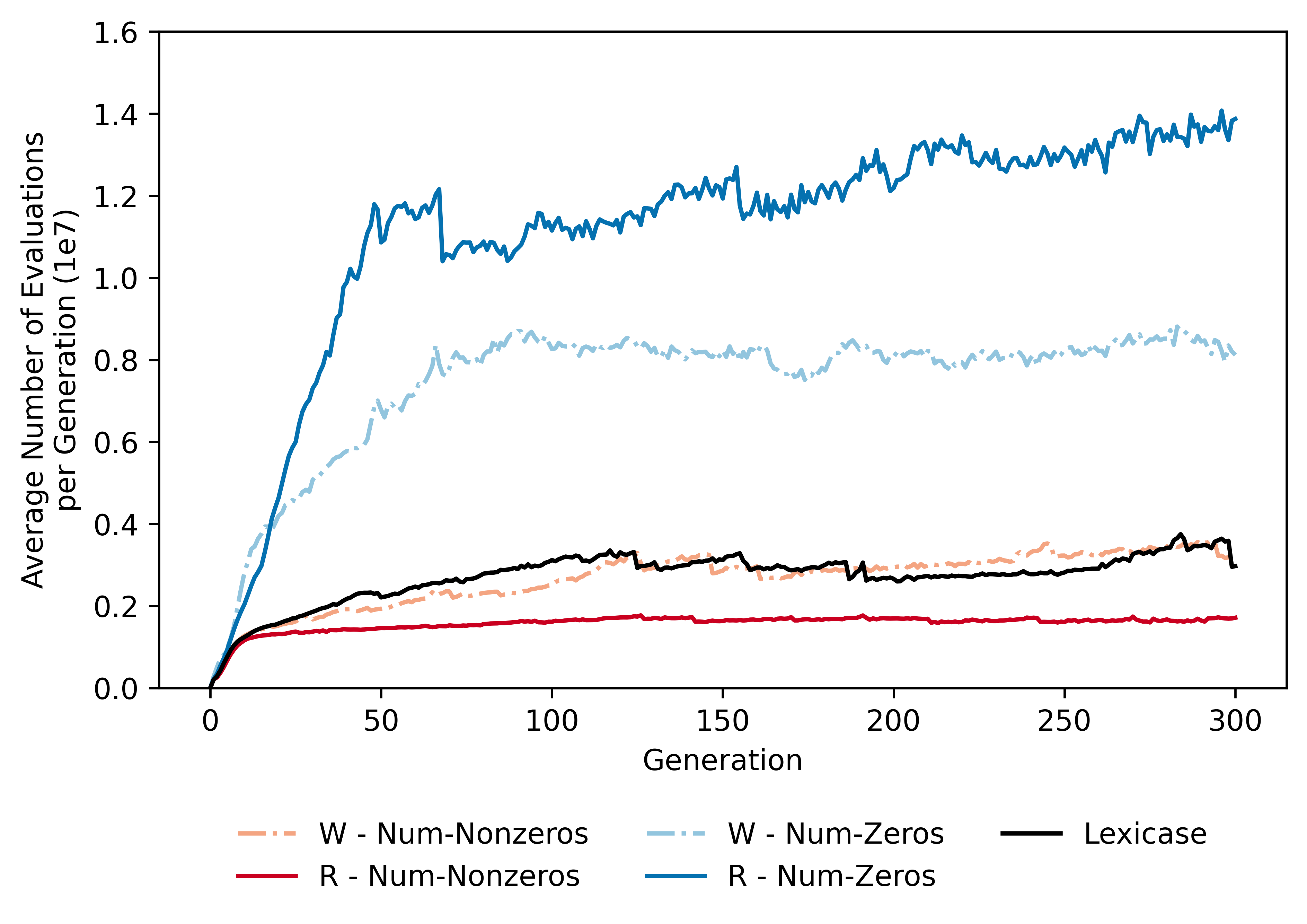}
    \caption{Average number of evaluations performed in a given active generation over evolutionary time while solving the \emph{Last Index of Zeros} problem.}
    \label{fig:LIOZ_COMP_NC}
\end{figure}

Figures~\ref{fig:MI_COMP_NC} and~\ref{fig:LIOZ_COMP_NC} show how the average number of comparisons per generation are affected by these different selection methods for the \emph{Mirror Image} and \emph{Last Index of Zero} problems, respectively. It is clear from both of these figures that ranked shuffle with Number-of-Nonzeros is consistently performing less evaluations per generation than any of the other methods.

These results show that there are significant improvements in efficiency when using a method based on weighted shuffling with a hard-first metric. We also find that using hard-first shuffling methods have comparable performance (in terms of success rate) to randomly shuffled lexicase selection in GP. This result leads us to believe that using weighted shuffle in other applications of lexicase selection, especially ones with computationally intensive evaluations, would potentially reduce the overall computational cost of lexicase selection, while maintaining the numerous benefits it provides. In order to verify this claim, we repeat the above experiments in a different, more computationally intensive, domain: image classification with deep neural networks.

% \begin{figure}
%\caption{Comparison of Efficiency of Selection Strategies for Symbolic %Regression}
%\includegraphics[width=0.5\textwidth]{2022_02_gecco_weighted/figures/SR_COMP_NC.png}
%\end{figure}

\section{Experiments on Image Classification}

In this section, we present the experimental results of testing the proposed fast gradient lexicase selection method on the image
classification problem, which is a popular problem that has been extensively
studied in computer vision and deep learning literature. We use a common network
architecture, VGG~\citep{simonyan15very}, and test our method on a benchmark
dataset, CIFAR-10~\citep{krizhevsky2009learning}. The dataset comprises of 32 $\times$ 32
pixel real-world RGB images of common objects (50,000 for training, 10,000 for
testing).

\subsection{Bias Metrics and Weight Assignment}

In this experiment, we consider the same two bias metrics as we did for the GP problems: Number-of-Zeros (Num-Zeros) and Number-of-Nonzeros (Num-Nonzeros). For the image classification problem, the model returns a probability distribution of all the classes for each input case. So, we take the one with largest probability as the discrete prediction output. The Num-Zeros metric $n_{zeros}$ counts the number of individuals in the current population that achieve zero error, \ie, make the correct prediction, on the given training case $i$. We therefore assign the according weight as $W[i]= n_{zeros} + 1$, with $+1$ to avoid zero weights. This metric can be interpreted as giving easier cases more weight, so they tend to appear earlier in the sequence of training cases. Alternatively, the Num-Nonzeros metric $n_{nonzeros}$ counts the number of individuals that achieve non-zero error, which, in this experiment is referred to making a wrong prediction. This metric, as opposed to $n_{zeros}$, gives more priority to hard cases.

\subsection{Implementation Details}

We follow the general experiment setups as described in \cite{ding2022optimizing, ding2021evolving}. The network architecture used in this work is VGG~\cite{simonyan15very}. We use a population size $p=4$ for both gradient lexicase selection and fast gradient lexicase selection. We set the total number of epochs as $200(p+1)$. 

\subsection{Results}

We evaluate the proposed fast gradient lexicase selection method against the original lexicase selection on the image classification problem. Analyses on both speed and accuracy are presented as follows.

\paragraph{\textbf{Speed.}}
First, we compare the number of evaluations performed among all the design choices of ways to initialize default weight and bias metrics. The count of evaluations performed in each generation can be viewed as the theoretical runtime of lexicase selection, under the assumption that a single evaluation step, \eg, inference of a DNN on a single input case, takes significantly more time than other operations such as comparing values. 

\begin{figure}
  \begin{center}
      \includegraphics[width=.99\linewidth]{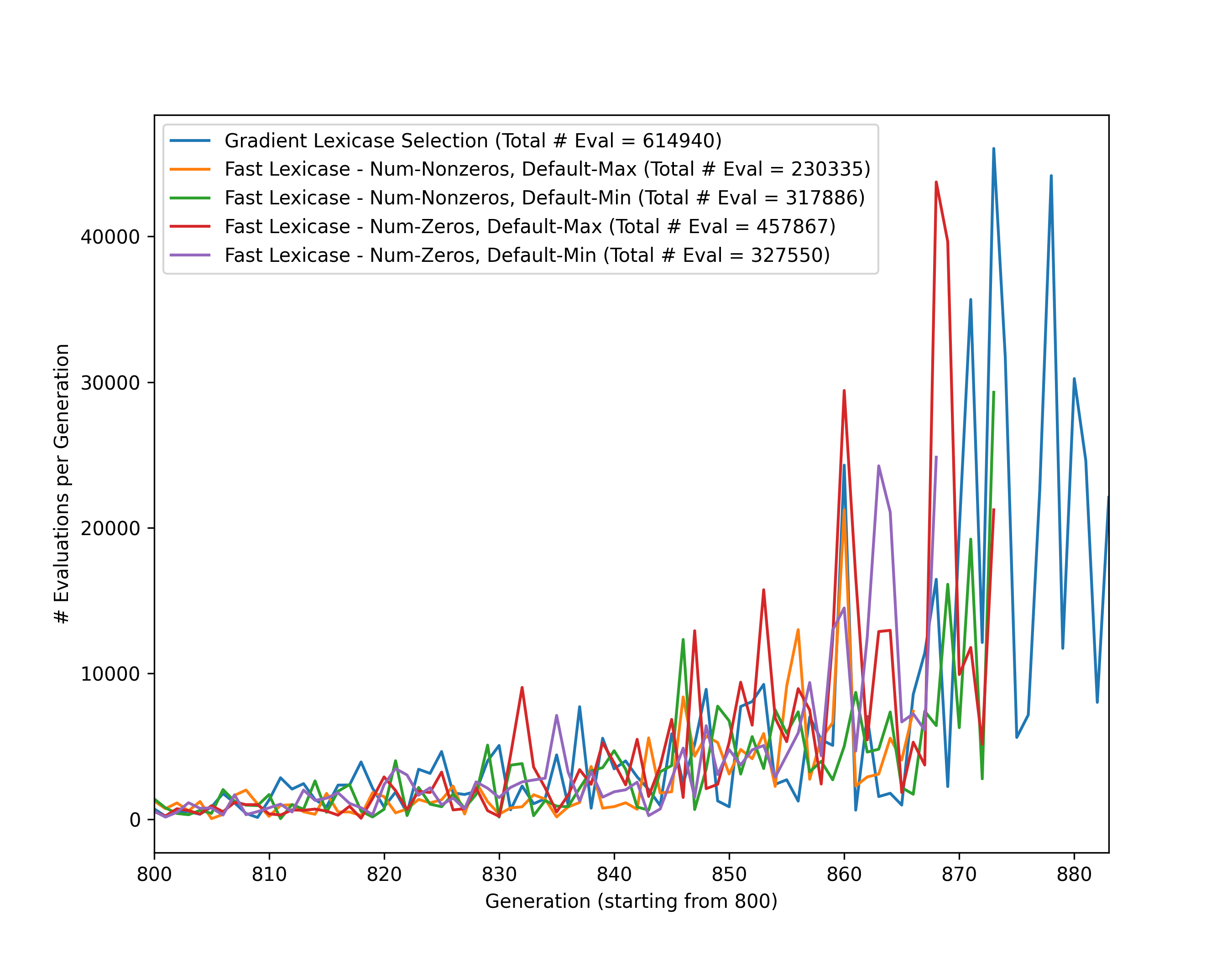}
  \end{center}
  \caption{Comparing number of evaluations of regular gradient lexicase selection and fast gradient lexicase selection (with different design choices). The plot skips early generations because most of those generations only require very few number of evaluations. We can see that the fast gradient lexicase selection methods in general requires significantly fewer evaluations than the regular gradient lexicase selection, and also requires fewer generations.}
  \label{fig:dl1}
\end{figure}

The raw evaluation counts per generation is shown in Fig.~\ref{fig:dl1}. We can see that in this problem, most of the evaluations happen in the last few generations. Fast gradient lexicase selection in general ends up with fewer generations for evolution, and requires fewer evaluations to be performed in each generation. 

\begin{figure}
  \begin{center}
      \includegraphics[width=.99\linewidth]{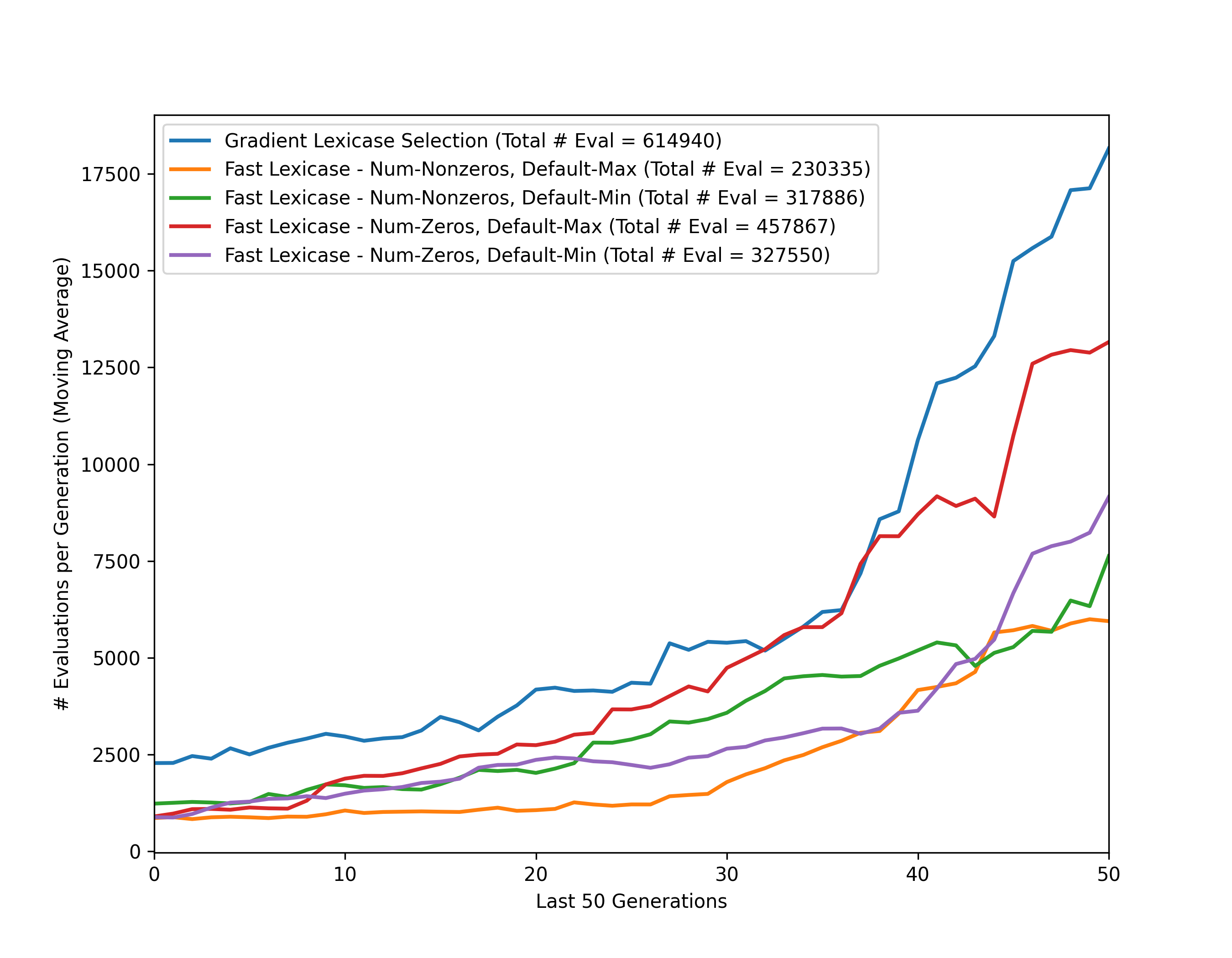}
  \end{center}
  \caption{Comparing number of evaluations of regular gradient lexicase selection and fast gradient lexicase selection, using the same data as Fig.~\ref{fig:dl1} but on a different view with the last 50 generations of each method, aligned by the end. The y-axis is smoothed by applying a moving average over 20 generations. This plot more clearly shows the comparison among methods, that the fast gradient lexicase selection with Default-Max and Num-Nonzeros being the most efficient method.  
  }
  \label{fig:dl2}
\end{figure}

To have a better view of the results, we create another view of Fig.~\ref{fig:dl1} with the last 50 generations of each method, aligned by the end. We focus on those later generations because early generations require many fewer evaluations, and thus does not contribute much to the total runtime. We also perform a moving average of 20 generations to smooth the values of y-axis. As shown in Fig.~\ref{fig:dl2}, we can see a clear comparison between the methods. The fast gradient lexicase selection with Default-Max and Num-Nonzeros requires significantly fewer evaluations throughout generations. These results aligns well with our findings for GP, as described in Sec.~\ref{sec:caseresult}

\paragraph{\textbf{Accuracy.}}
While the results show that fast gradient lexicase selection is significantly
more efficient than the regular method, we also need to test whether the reduced
number of evaluations has an effect on the performance on the image
classification task. Table~\ref{tab:dl1} shows quantitative results of model
performance. We can see that the performance of all the methods are similar. To
statistically illustrate this, for each of the fast gradient lexicase selection
methods, we perform the one proportion z-test against the regular gradient
lexicase selection. 

Let $p$ be the accuracy of tested method (fast gradient lexicase selection) and $p_0$ be the hypothesized accuracy of the baseline method (gradient lexicase selection). For the z-test, we use the following null hypotheses:
\[
  H_0: p=p_0
\]
and the left-tailed alternative hypothesis:
\[
  H_1: p<p_0
\]

If the p-value that corresponds to the test statistic is less than the
significance level, we can reject the null hypothesis. From Table~\ref{tab:dl1},
we can see that if we choose the common significance level $0.05$, none of the
methods have a p-value less than $0.05$. So, we conclude that the performance of
fast gradient lexicase selection (with any design choice considered in this
work) is not significantly different from the performance of regular gradient
lexicase selection.

\section{Discussion}

When using lexicase selection, some cases have a large impact on how many individuals are filtered out of the selection pool, while other cases have little to no impact. Additionally, some cases are \textit{synonymous} (or nearly synonymous), in that they test similar inputs and require very similar behavior of individuals to perform well on. While one synonymous case may filter the selection pool, consequent cases may have little effect.

Our primary motivation of using weighted shuffle for lexicase selection is to bias the shuffle such that cases that have more impact on the filtering of the selection pool more often occur near the start of the shuffled list of cases. By doing so, the selection pool size will shrink more rapidly, requiring fewer evaluations during selection. Before this work, it was an open question whether certain shuffling techniques or bias metrics would result in more rapid filtering of the selection pool, and if so, whether they would also have any effect on problem-solving performance.

Our results, both using PushGP for program synthesis and deep neural networks for image classification, indicate that weighted shuffle does have a positive impact on running time while not negatively affecting problem-solving performance compared to uniform shuffle. In particular, methods that weight cases that are more difficult to answer correctly more heavily (Num-Nonzeros) used the fewest comparisons or evaluations, and therefore better running time.

More difficult cases are likely passed by fewer individuals in the population than easier cases. When putting more emphasis on these cases by having them biased to occur earlier in the weighted shuffle, they likely remove more individuals from the selection pool, keeping around the smaller group of individuals who perform better on them. However, this result is not obvious \textit{a priori}; it might have been the case that all individuals perform equally poorly on some difficult cases, meaning that they apply less filtering power on the selection pool, and that putting easier cases first (or a uniform shuffle) would produce faster filtering.

\begin{table}[t]
  \caption{Image classification results. We report the percentage accuracy (acc.) for all the methods. For each of the fast gradient lexicase selection methods, we also report the p-value of one proportion z-test against the regular gradient lexicase selection. A p-value greater than the significance level (0.05) means there is no significance different between the two methods in performance.}
  \label{tab:dl1}
  \begin{center}
  \begin{tabular}{lcc}
  \toprule
  \multicolumn{1}{c}{\bf Method}  &\multicolumn{1}{c}{\bf acc.} &\multicolumn{1}{c}{\bf p-val}\\
  \midrule 
  Gradient Lexicase Selection & 93.34 & -	\\
  Fast Lexicase  - Nonzeros, Default-Max & 93.19 & 0.276\\
  Fast  Lexicase  - Nonzeros, Default-Min & 92.99 & 0.085\\
  Fast  Lexicase  - Zeros, Default-Max & 93.29 & 0.421\\
  Fast  Lexicase  - Zeros, Default-Min & 93.17	& 0.25\\
  \bottomrule
  
\end{tabular}
\end{center}
\end{table}

Even though biasing easier cases earlier (Num-Zeros) did not reduce the number of evaluations of gradient lexicase selection as much biasing harder cases first, it did nevertheless produce significantly fewer evaluations than uniform shuffle in our deep learning experiments. While not intuitive, we suppose that both easier and harder cases may result in increased filtering pressure compared to average cases. A possible explanation is that the benefit of weighted shuffling may actually come from the behavior of focusing on some set of cases (no matter easy or hard ones), which leads to faster evolution compared to looking at random cases.
% \todo{Ding}{I don't have a great reason here, any intuitions?} -- I added a few.

We also find that the initialization of case weights is important to consider when performing fast gradient lexicase selection. When using a metric that places higher weight on hard cases, it seems to be beneficial to initialize all the cases to be maximally difficult with Default-Max. This might be because there is an initial bias towards looking at cases that have not been seen yet (as these are set as being hard) and therefore leads to a better initial exploration of the case space. As we iteratively assign weights (that might be less than or equal to the max), the probability we select these seen cases becomes less than that for an unseen (maximally hard) case. We find that this explanation is also supported by the easy first metric. When using a shuffling method that places emphasis on easy cases, such as Number-of-Zeros, initializing cases to be minimally difficult would place an initial bias towards looking at cases that have not been considered yet. This might explain why Default-Max outperforms Default-Min for the Number-of-Nonzeros bias metric, and the opposite for the Number-of-Zeros metric in our fast gradient lexicase selection experiments.

\section{Conclusion and Future Work}

In this work, we demonstrate the merit of using weighted shuffle to increase the efficiency of the selection procedure. In order to overcome the efficiency disadvantage of lexicase selection for large-scale evolutionary computation, we propose fast lexicase selection, which is an attempt to combine weighted shuffle and online weight updating with lexicase selection. The experimental results on classic GP problems show that there are significant improvements in efficiency when using the hard-first weighted shuffling methods, and the performance is still comparable to regular lexicase selection. 

We further extend the idea to the gradient lexicase selection method, which can better demonstrate the enhanced efficiency of fast lexicase selection in computationally intensive domains. We test our method on a large-scale image classification problem. Experimental results show that fast lexicase selection with different weighting methods outperforms regular lexicase selection in terms of efficiency, without hindering accuracy. More specifically, the hard-first metric with the initialization of cases to be maximally hard gives the best efficiency. We also find that when using the fast lexicase selection method, initializing cases as being maximally hard when using a hard-first metric, or minimally hard when using an easy-first metric seems to be beneficial in decreasing the number of evaluations needed for every generation. 

Through this work, we highlight the importance of using the information about how the population performs on training cases in the selection procedure to identify cases with higher selection pressure, and thus improve the efficiency of the selection procedure. This is especially important in large scale evolutionary optimization problems, where each evaluation step might be costly, or the total number of evaluations might be large. For future work, we look forward to the exploration of better ways to extract the incremental information about the relationship between population and training data, in order to improve the algorithm in terms of both efficiency and performance.

\begin{acks}

This work is supported by the National Science Foundation
under Grant No. 1617087. Any opinions, findings, and conclusions expressed in this publication are those of the authors and do not necessarily reflect the views of the National Science Foundation. 

This work was performed in part using high performance computing equipment obtained from the Collaborative R\&D Fund managed by the Massachusetts Technology Collaborative. 

The authors would like to thank Edward Pantridge and Anil Saini for their valuable comments and helpful suggestions.

\end{acks}

\bibliographystyle{ACM-Reference-Format}
\bibliography{bibcombined}

\end{document}